\documentclass[11pt,a4paper]{article}
\usepackage[hyperref]{emnlp-ijcnlp-2019}
\usepackage{times}
\usepackage{latexsym}

\usepackage{url}

\usepackage{array}
\newcommand{\PreserveBackslash}[1]{\let\temp=\\#1\let\\=\temp}
\newcolumntype{C}[1]{>{\PreserveBackslash\centering}p{#1}}
\newcolumntype{R}[1]{>{\PreserveBackslash\raggedleft}p{#1}}
\newcolumntype{L}[1]{>{\PreserveBackslash\raggedright}p{#1}}

\usepackage{amsfonts}
\usepackage{graphicx}
\usepackage{subfigure}
\usepackage{multirow}
\usepackage{arydshln}
\usepackage[english]{babel}

\aclfinalcopy 

\title{Cross-Lingual BERT Transformation for Zero-Shot Dependency Parsing}

\author{Yuxuan Wang$^1$, Wanxiang Che$^1$\thanks{~~Email corresponding}, Jiang Guo$^2$, Yijia Liu$^1$, and Ting Liu$^1$ \\
	$^1$Research Center for Social Computing and Information Retrieval, Harbin Institute of Technology \\
	$^2$Computer Science and Artificial Intelligence Laboratory, MIT \\
	{\tt \{yxwang, car, yjliu, tliu\}@ir.hit.edu.cn, jiang\_guo@csail.mit.edu}
}

\date{}

\begin{document}
\maketitle
\begin{abstract}

This paper investigates the problem of learning cross-lingual representations in a contextual space. We propose \textbf{C}ross-\textbf{L}ingual \textbf{B}ERT \textbf{T}ransformation (CLBT), a simple and efficient approach to generate cross-lingual contextualized word embeddings based on publicly available pre-trained BERT models \cite{devlin2018bert}.
In this approach, a linear transformation is learned from contextual word alignments to align the contextualized embeddings independently trained in different languages.
We demonstrate the effectiveness of this approach on zero-shot cross-lingual transfer parsing. 
Experiments show that our embeddings substantially outperform the previous state-of-the-art that uses static embeddings. We further compare our approach with XLM \cite{lample2019cross}, a recently proposed cross-lingual language model trained with massive parallel data, and achieve highly competitive results. \footnote{Our code is released at \url{https://github.com/WangYuxuan93/CLBT}}
\end{abstract}

\section{Introduction}

One of the most promising directions for cross-lingual dependency parsing, which also remains a challenge, is to bridge the gap of lexical features. 
Prior works \cite{W14-1613, guo-EtAl:2015:ACL-IJCNLP2} have shown that cross-lingual word embeddings are able to significantly improve the transfer performance compared to delexicalized models \cite{mcdonald2011multi, mcdonald2013universal}. 
These cross-lingual word embeddings are \textit{static} in the sense that they do not change with the context.\footnote{In this paper, we refer to these embeddings as \textit{static} as opposed to \textit{contextualized} ones.} 

Recently, contextualized word embeddings derived from large-scale pre-trained language models \cite{NIPS2017_7209,peters2017semi,peters2018deep,devlin2018bert} have demonstrated dramatic superiority over traditional static word embeddings, establishing new state-of-the-arts in various monolingual NLP tasks \cite{suzana2018deep, schuster2018cross}.
The success has also been recognized in dependency parsing \cite{che2018towards}.
The great potential of these contextualized embeddings has inspired us to extend its power to cross-lingual scenarios.


\begin{figure}
	\centering
	\includegraphics[width=75mm]{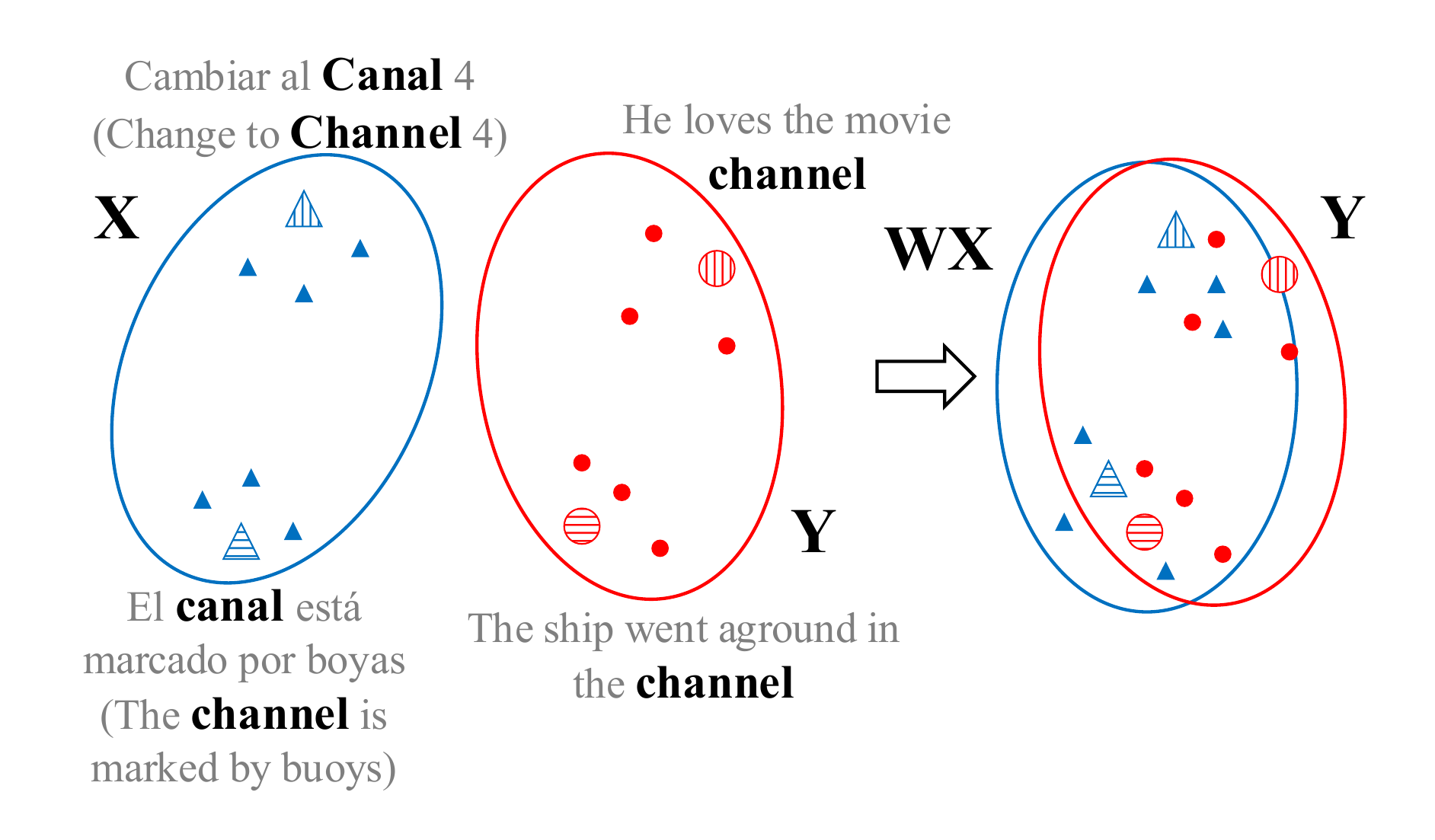}
	\caption{A toy illustration of the method, where contextualized embeddings of the word \textit{canal} from Spanish is transformed to the semantic space of English.}
	\label{fig:toy-illustration}
\end{figure}

Several recent works have been proposed to learn contextualized cross-lingual embeddings by training cross-lingual language models from scratch with parallel data as supervision, and has been demonstrated effective in several downstream tasks \cite{schuster2018cross, mulcaire2019polyglot, lample2019cross}.
These methods are typically resource-demanding and time-consuming.\footnote{For instance, XLM was trained on 64 Volta GPUs \cite{lample2019cross}. While the time of training is not described in the paper, we may take the statistics from BERT as a reference, e.g., BERT$_{\rm {BASE}}$ was trained on 4 Cloud TPUs for 4 days \cite{devlin2018bert}.}
In this paper, we propose \textbf{C}ross-\textbf{L}ingual \textbf{B}ERT \textbf{T}ransformation (CLBT), a simple and efficient off-line approach that learns a linear transformation from contextual word alignments. 
With CLBT, contextualized embeddings from pre-trained BERT models in different languages are projected into a shared semantic space.
The learned transformation is then used on top of the BERT encodings for each sentence, which are further fed as input to a parser.

Our approach utilizes the semantic equivalence in word alignments, and thus is supposed to be \textit{word sense-preserving}. Figure~\ref{fig:toy-illustration} illustrates our approach, where contextualized embeddings of the Spanish word ``canal" are transformed to the corresponding semantic space in English.

Experiments on the Universal Dependencies (UD) treebanks (v2.2) 
\cite{nivre2018ud} show that our approach substantially outperforms previous models that use static cross-lingual embeddings, with an absolute gain of 2.91\% in averaged LAS.
We further compare to XLM \cite{lample2019cross}, a recently proposed large-scale cross-lingual language model. Results demonstrate that our approach requires significantly fewer training data, computing resources and less training time than XLM, yet achieving highly competitive results.

\section{Related Work}

Static cross-lingual embedding learning methods can be roughly categorized as \textit{on-line} and \textit{off-line} methods. Typically, on-line approaches integrate monolingual and cross-lingual objectives to learn cross-lingual word embeddings in a joint manner \cite{C12-1089, P14-2037, guo2016representation}, while off-line approaches take pre-trained monolingual word embeddings of different languages as input and retrofit them into a shared semantic space \cite{xing2015normalized,lample2018word,chen2018unsupervised}.

Several approaches have been proposed recently to connect the rich expressiveness of contextualized word embeddings with cross-lingual transfer. 
\newcite{mulcaire2019polyglot} based their model on ELMo \cite{peters2018deep} and proposed a polyglot contextual representation model by capturing character-level information from multilingual data.
\newcite{lample2019cross} adapted the objectives of BERT \cite{devlin2018bert} to incorporate cross-lingual supervision from parallel data to learn cross-lingual language models (XLMs), which have obtained state-of-the-art results on several cross-lingual tasks.
Similar to our approach, \citet{schuster2019cross} also aligned pre-trained contextualized word embeddings through linear transformation in an off-line fashion. They used the averaged contextualized embeddings as an \textit{anchor} for each word type, and learn a transformation in the anchor space. Our approach, however, learns this transformation directly in the contextual space, and hence is explicitly designed to be \textit{word sense-preserving}.


\section{Cross-Lingual BERT Transformation}

This section describes our proposed approach, namely CLBT, to transform pre-trained monolingual contextualized embeddings to a shared semantic space.

\subsection{Contextual Word Alignment}

Traditional methods of learning static cross-lingual word embeddings have been relying on various sources of supervision such as bilingual dictionaries \cite{lazaridou2015hubness, smith2017offline}, parallel corpus \cite{guo-EtAl:2015:ACL-IJCNLP2} or on-line Google Translate \cite{mikolov2013exploiting,xing2015normalized}.
To learn contextualized cross-lingual word embeddings, however, we require supervision at word token-level (or context-level) rather than type-level (i.e. dictionaries). Therefore, we assume a parallel corpus as our supervision, analogous to on-line methods such as XLM \cite{lample2019cross}.

In our approach, unsupervised bidirectional word alignment is first applied to the parallel corpus to obtain a set of aligned word pairs with their contexts, or \textit{contextual word pairs} for short.
For one-to-many and many-to-one alignments, we use the left-most aligned word,\footnote{Preliminary experiments indicate that this way works better than keeping all the alignments.} such that all the resulting word pairs are one-to-one.
In practice, since WordPiece embeddings \cite{wu2016google} are used in BERT, all the parallel sentences are tokenized using BERT's wordpiece vocabulary before being aligned.

\subsection{Off-Line Transformation}

Given a set of contextual word pairs, their BERT representations $\{\mathbf{x}_i,\mathbf{y}_i\}_{i=1}^n$ can be easily obtained from pre-trained BERT models,\footnote{In this work, we use the English BERT (enBERT) for the source language (English) and the multilingual BERT (mBERT), which is trained on 102 languages without cross-lingual supervision, for all the target languages.} where $\mathbf{x}_i \in \mathbb{R}^{d_1} $ is the contextualized embedding of token $i$ in the target language, and $\mathbf{y}_i \in \mathbb{R}^{d_2} $ is the representation of its alignment in the source language.

In our experiments, a parser is trained on source language data and applied directly to all the target languages. Therefore, we propose to project the embeddings of target languages to the space of the source language, instead of the opposite direction.
Specifically, we aim at finding an appropriate linear transformation $\mathbf{W}$, such that $\mathbf{Wx}_i$ approximates $\mathbf{y}_i$.\footnote{We also investigated non-linear transformation in our experiments, but didn't observe any improvements.}
This can be achieved by solving the following optimization problem:
$$\min_\mathbf{W} \sum_{i=1}^{n} ||\mathbf{Wx}_i-\mathbf{y}_i||^2 ,$$
where $\mathbf{W} \in \mathbb{R}^{d_1 \times d_2} $ is a parameter matrix.

Previous works on static cross-lingual embeddings have shown that an orthogonal $\mathbf{W}$ (i.e. $\mathbf{W}^\top \mathbf{W}=\mathbf{I}$) is helpful for the word translation task \cite{xing2015normalized}.
In this case, an analytical solution can be found through singular value decomposition (SVD) of $\mathbf{Y}^\top \mathbf{X}$:
$$ \mathbf{W} = \mathbf{VU}^\top \textrm{, where } \mathbf{U\Sigma V}^\top = \textrm{SVD} (\mathbf{Y}^\top \mathbf{X}). $$

Here $\mathbf{X} \in \mathbb{R}^{n \times d}$ and $\mathbf{Y} \in \mathbb{R}^{n \times d}$ are the contextualized embedding matrices, where $n$ is the number of aligned contextual word pairs, $d$ is the dimension of monolingual contextualized embeddings. Each pair of rows $(\mathbf{x}_i, \mathbf{y}_i)$ indicates an aligned contextual word pair.

Although this can be computed in CPUs within several minutes, more memories will be required with the growth of the amount of training data.
Therefore, we present an approximate solution, where $\mathbf{W}$ is optimized with gradient decent (GD) and is not constrained to be orthogonal.\footnote{We found the orthogonal constraint doesn't help for GD.}
This GD-based approach can be trained on a single GPU and typically converges in several hours.

To validate the effectiveness of our approach in cross-lingual dependency parsing, we first obtain the CLBT embeddings with the proposed approach, and then use them as input to a modern graph-based neural parser (described in next section), in replacement of the pre-trained static embeddings.
Note that BERT produces embeddings in wordpiece-level, so we only use the left-most wordpiece embedding of each word.\footnote{We tried alternative strategies such as averaging, using the middle or right-most wordpiece, but observed no significant difference.} 

\section{Experiments}
\label{sec:experiments}
\subsection{Data and Settings}

In our experiments, the contextual word pairs are obtained from the Europarl corpora \cite{koehn2005epc} using the \textit{fast\_align} toolkit \cite{dyer2010cdec}. 
Only 10,000 sentence pairs are used for each target language.
For the parsing datasets, we use the Universal Dependencies(UD) Treebanks (v2.2) \cite{nivre2018ud},\footnote{\url{hdl.handle.net/11234/1-2837}} following the settings of the previous state-of-the-art system \cite{ahmad2018near}.
From the 31 languages they have analyzed, we select 18 whose Europarl data is publicly available.\footnote{For languages with multiple treebanks, we use the same combinations as they did.}  
Statistics of the selected languages and treebanks can be found in the Appendix.
We employ the Biaffine Graph-based Parser of \newcite{dozat2017deep} and adopt their hyper-parameters for all of our models.

In all the experiments, English is used as the source language, and the other 17 languages as targets. 
The model is trained on the English treebank 
and applied directly to target languages with the transformed contextualized embeddings. 
We train our models using the Adam optimizer \cite{kingma2015adam}, and most of the them converge within a few thousand epochs in several hours. More implementation details are reported in the Appendix.

\subsection{Baseline Systems}

We compare our method with the following three baseline models:

\begin{itemize}
    \vspace{-0.5em}
	\item \textbf{mBERT} (contextualized). Embeddings generated by the mBERT model are directly used in the training and testing procedures.
	\vspace{-0.5em}
	\item \textbf{FT-SVD} \cite[off-line, static]{ahmad2018near}. SVD-based transformation \cite{smith2017offline} is applied on 300-dimensional FastText embeddings \cite{bojanowski2017enriching} to obtain cross-lingual static embeddings, which represents the previous state-of-the-art.
	 We report results from their paper of the RNNGraph model which used the same architecture as ours.
	\vspace{-0.5em}
	\item \textbf{XLM} \cite[on-line, contextualized]{lample2019cross}. A strong method which learns contextualized cross-lingual embeddings from scratch with cross-lingual data. 
\end{itemize}

For the \textbf{XLM} model, we employ the XNLI-15 model\footnote{\url{github.com/facebookresearch/XLM}} they released to generate embeddings and apply them to cross-lingual dependency parsing in the same way as we do with our own model.
We compare with them in the 4 overlapped languages both works have researched on.


\subsection{Comparison with Off-Line Methods}

\begin{table}[t]
	\centering
	\small
	\renewcommand{\arraystretch}{1.2}
	\begin{tabular}{l|l|l||l|l}
		\hline
		\multirow{3}{*}{\bf Lan.}&Static & \multicolumn{3}{c}{Contextualized} \\
		\cline{2-5}
		&\multirow{2}{1.2cm}{\bf FT-SVD} 
		&\multirow{2}{1.1cm}{\bf mBERT} 
		&\multirow{2}{1cm}{\bf CLBT (SVD)} 
		&\multirow{2}{1cm}{\bf CLBT (GD)} \\
		& & & & \\
		\hline
		en &88.31 &90.71 &\multicolumn{2}{c}{\bf 91.03*} \\
		\hdashline
		\hdashline
		de &59.31 &63.41 &\bf 64.47* &62.14 \\
		da &68.81 &70.57 &71.60* &\bf 71.66* \\
		sv &73.49 &70.09 &73.33* &\bf 75.95* \\
		nl &60.11 &\bf 65.66 &65.45 &63.86 \\
		\hdashline
		fr &73.46 &72.97 &74.70* &\bf 76.59* \\
		it &76.23 &79.02 &\bf 79.46 &78.98 \\
		es &66.91 &65.43 &67.14* &\bf 68.33* \\
		pt &67.98 &67.11 &69.12* &\bf 69.25* \\
		ro &52.11 &46.40 &55.14* &\bf 55.84* \\
		\hdashline
		sk &56.98 &50.76 &59.46* &\bf 59.92* \\
		pl &58.59 &63.10 &65.37* &\bf 65.80* \\
		bg &66.68 &\bf 71.20 &70.26 &70.75 \\
		sl &54.57 &56.78 &\bf 57.42* &57.21* \\
		cs &52.80 &45.20 &52.20* &\bf 52.99* \\
		\hdashline
		fi &48.74 &49.56 &51.00* &\bf 52.61* \\
		et &44.40 &46.64 &47.79* &\bf 48.52* \\
		\hdashline
		lv &49.59 &45.11 &48.59* &\bf 49.78* \\
		\hline
		AVG. &60.63 &60.53 &63.09 &\bf 63.54 \\
		\hline
	\end{tabular}
	\caption{Results (LAS\%) on test sets. Languages are split by language families with dashed lines. AVG. means the average of results from all target languages. Statistically significant differences between our methods and the mBERT model are marked with *, with p-value \textless ~0.05 under McNemar's test.} 
	\label{tbl:main-results}
	\vspace{-1em}
\end{table}

Results on the test sets are shown in Table~\ref{tbl:main-results}.\footnote{UAS results are listed in the Appendix due to space limit. Note that since we have no access to the parsed files of the FT-SVD model, we only report statistical significant tests between our methods and the mBERT model, which is highly comparable to the FT-SVD model on average.} Languages are grouped by language families. 
Overall, our approach with either SVD or GD outperforms both FT-SVD and mBERT by a substantial margin (+2.91\% in averaged LAS), among which GD turns out to be slightly better than SVD in most of the languages. 
When combined with FT-SVD, the performances can be further improved by 0.33\% in LAS for the GD method and 0.51\% for SVD (see the Appendix for more details). 
Interestingly, the mBERT model which is trained without any cross-lingual supervision but using a shared multilingual wordpiece vocabulary works surprisingly well in some languages, especially in those linguistically close to English. Similar observations have also been identified in other works \cite{pires2019multilingual,wu2019beto}.

\begin{table}[tbp]
	\centering
	\small
	\renewcommand{\arraystretch}{1.2}
	\begin{tabular}{l|l||l|l}
		\hline
		\bf Lan. &\bf XLM &\bf CLBT (SVD) &\bf CLBT (GD) \\
		\hline
		en &91.85/89.92 &\multicolumn{2}{c}{\bf 92.81*/91.03*} \\
		\hdashline
		de &\bf 74.65/65.31 &73.72/64.47 &71.08/62.14 \\
		fr &79.62/73.41 &80.01/74.70* &\bf 80.85*/76.59* \\
		es &75.41/67.43 &75.52/67.14* &\bf 75.70*/68.33* \\
		bg &81.07/69.45 &\textbf{82.14*}/70.26 &81.51/\textbf{70.75*} \\
		\hline
		AVG. &77.69/68.90 &\textbf{77.85}/69.14 &77.29/\textbf{69.45} \\
		\hline\hline
		\bf Data &0.2-13.1M &\multicolumn{2}{c}{10K} \\
		\hline
	\end{tabular}
	\caption{Results (UAS\%/LAS\%) on test sets. The last row shows the training data used in each language by sentence. AVG. means the average of results from 4 target languages. Statistically significant differences between our methods and the XLM are marked with an asterisk, with p-value \textless ~0.05 under McNemar's test.}
	\label{tbl:vs-xlm}
\end{table}

\subsection{Comparison with On-Line Methods}

Comparison of our approach and a cross-lingual language model pre-training (XLM) method \cite{lample2019cross} in the 4 overlapped languages is shown in Table~\ref{tbl:vs-xlm}. 
CLBT outperforms XLM in 3 out of the 4 languages but lower in German (de).
The amount of training data used in each method is also shown in the bottom: the number of parallel sentences used by XLM ranges from 0.2 million (10 million tokens) for Bulgarian to 13.1 million (682 million tokens) for French.
In comparison, only 10,000 parallel sentences (0.4 million tokens) are used for each language in CLBT, demonstrating the data-efficiency of our approach.
Moreover, given the efficiency in both data and training, CLBT can be readily scaled to new language pairs in hours.


\subsection{Analysis}

\subsubsection{Transformation of Cross-lingual BERT Embedding}
\label{sec:word-sense-alignment}


In order to investigate the properties of contextualized representations before and after the linear transformation, we employ the SENSEVAL2 data \cite{edmonds2001senseval2},\footnote{\url{www.hipposmond.com/senseval2/}} where words from different languages are tagged by their word senses in different contexts.

\begin{figure}[t]
	
	\centering
	\subfigure[]{
		\label{fig:en-es-before}
		\centering
		\includegraphics[width=60mm]{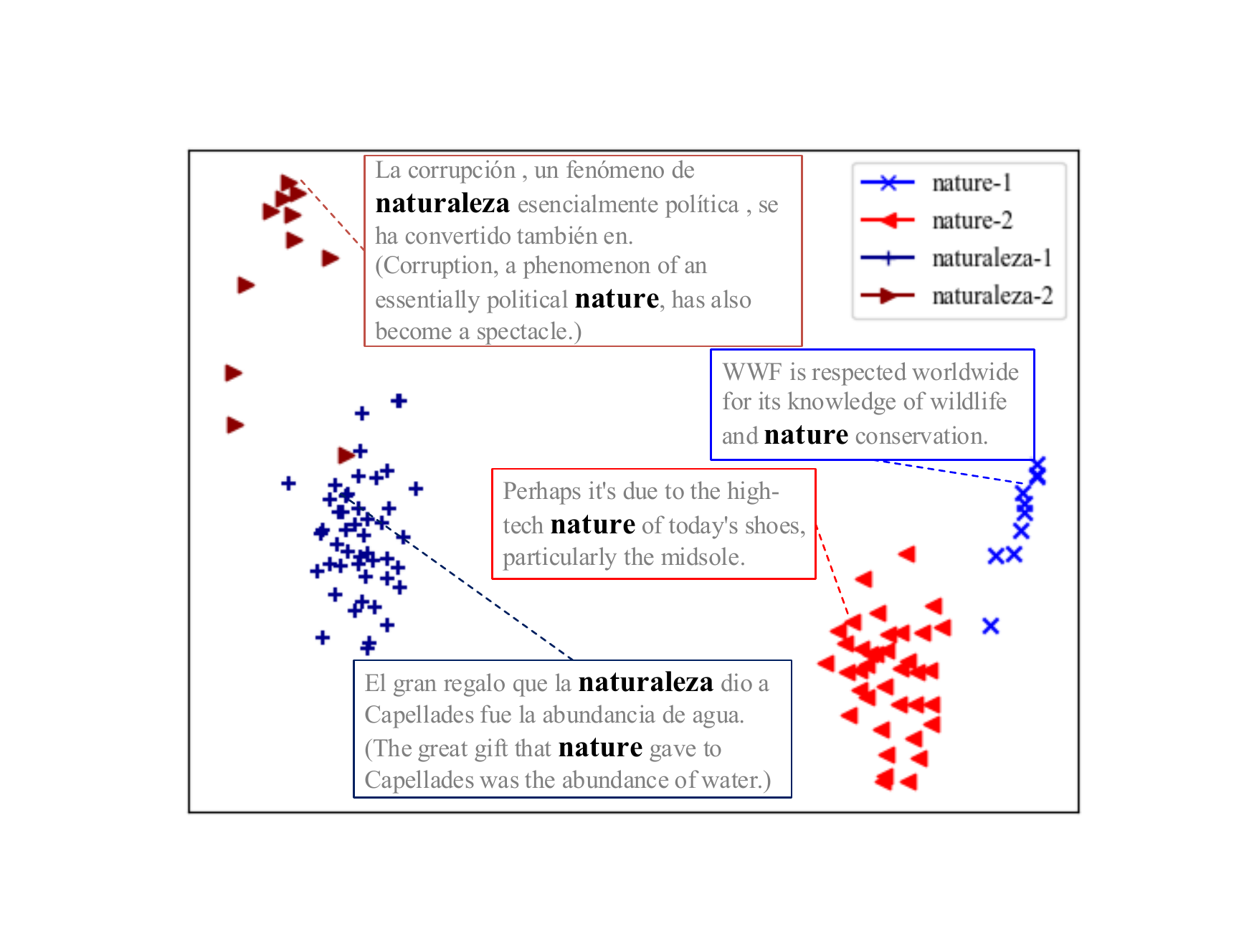}
	}
	\subfigure[]{
		\label{fig:en-es-after}
		\centering
		\includegraphics[width=60mm]{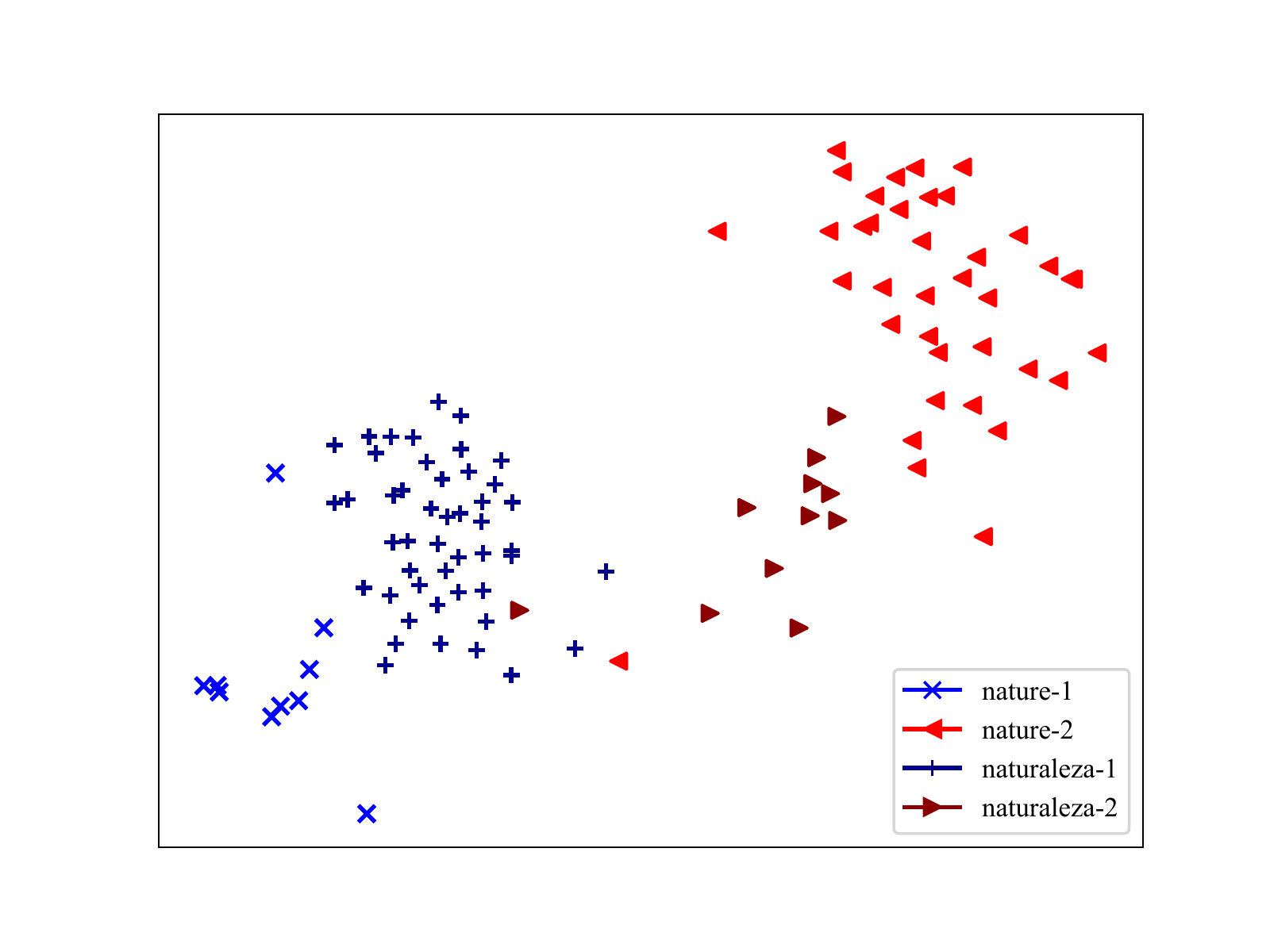}
	}
	\vspace{-0.5em}
	\caption{t-SNE visualization of the English word \textit{nature} and its Spanish translation \textit{naturaleza} in different contexts by the contextualized representations before (a) and after (b) the linear transformation. Points are colored by word senses. Example contexts are given in (a). Translations of Spanish sentences are in brackets.}
\end{figure}
	
We took contextualized representations of the English word \textit{nature} and its Spanish translation \textit{naturaleza} in different contexts from pre-trained English and multilingual BERT respectively and visualize their distributions in Figure~\ref{fig:en-es-before}, where we can observe obvious clustering of word senses. 
Specifically, words with sense \textit{nature-1} and \textit{naturaleza-1} mean \textit{the physical world}, whereas \textit{nature-2} and \textit{naturaleza-2} mean \textit{inherent features}.
We then apply our GD-based method to embeddings of \textit{naturaleza} and depict the resulting cross-lingual embeddings in Figure~\ref{fig:en-es-after}.
The distance between embeddings from English and Spanish is effectively reduced after the transformation. 
And it is apparent that embeddings of Spanish words are closer to those with similar meanings from English, which indicates the effectiveness of our approach.

\subsubsection{Effect of Training Data Size}
\label{sec:effect-of-amount}

We select several languages from each language family, and investigate the effect of the amount of training data on the performances of zero-shot cross-lingual dependency parsing.
Specifically, we take the SVD-based approach, since it is faster than the GD-based one, and trained different transformation models with different amount of parallel sentences from Europarl dataset on each of the 13 selected languages.

\begin{figure}
	\centering
	\includegraphics[width=65mm]{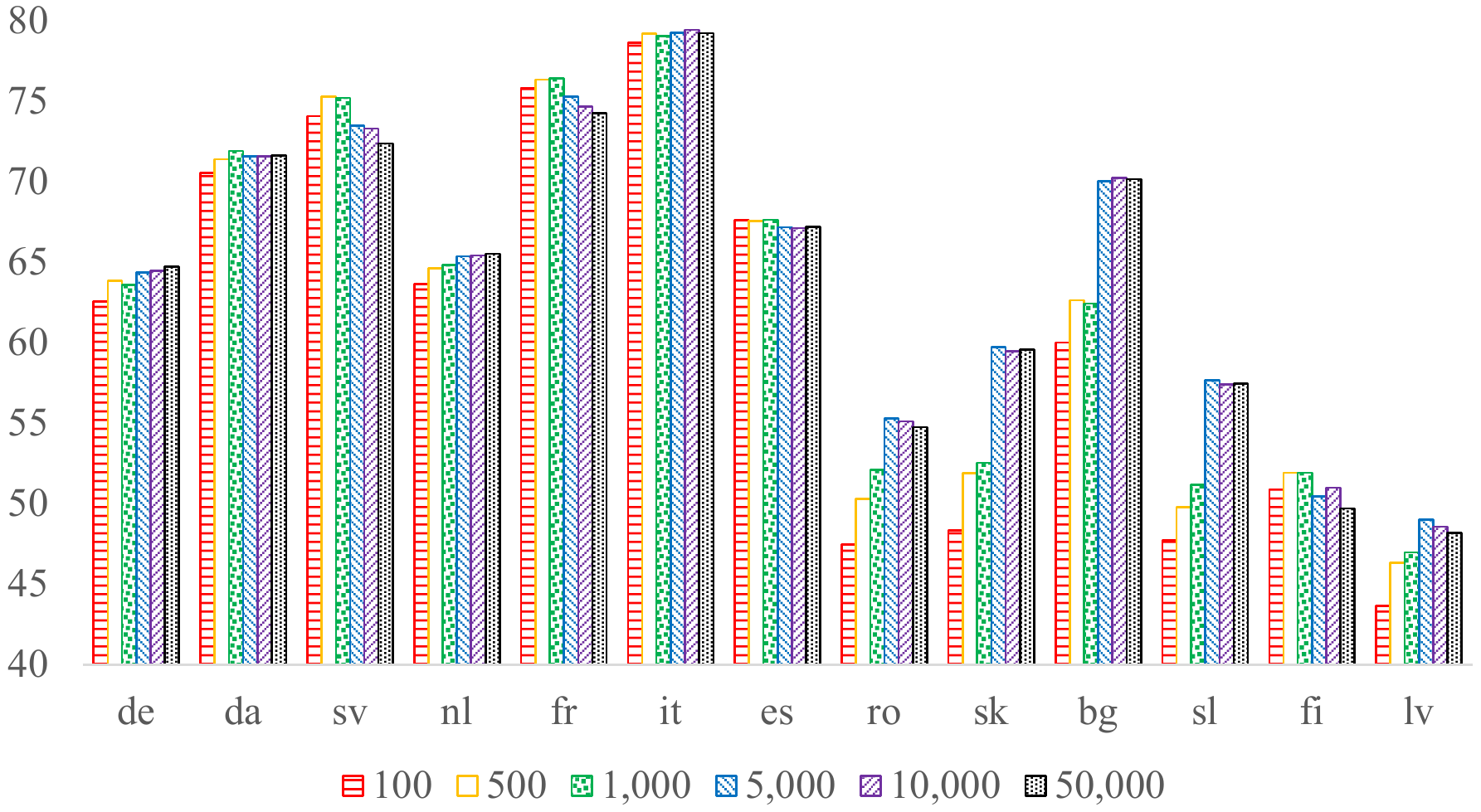}
	\caption{Effects of the amount of training data on different languages. ($y$-axis represents the LAS.)}
	\label{fig:svd-data-quantity}
\end{figure}

As shown in Figure~\ref{fig:svd-data-quantity}, for most of the languages, the best performance is achieved with only 5000 parallel sentences. 
It is also worth noting that for most of Germanic (e.g. German, Danish, Swedish and Dutch) and Romance (e.g. French, Italian, Spanish and Romanian) languages, which are typologically closer to English, a rather small training set of merely 100 sentences is capable of achieving comparative results.

\section{Conclusion}

We propose the Cross-Lingual BERT Transformation (CLBT) approach for contextualized cross-lingual embedding learning, which substantially outperforms the previous state-of-the-art in zero-shot cross-lingual dependency parsing.
By exploiting publicly available pre-trained BERT models, our approach provides a fast and data-efficient solution to learning cross-lingual contextualized embeddings.
Compared to the XLM, our method requires much fewer parallel data and less training time, yet achieving comparable performance. 

For future work, we are interested in unsupervised cross-lingual alignment, inspired by prior success on static embeddings \cite{lample2018word,alvarez2018gromov}, which demands a deeper understanding to the geometry of the multilingual contextualized embedding space.

\section*{Acknowledgments}
We thank the anonymous reviewers for their valuable suggestions. 
This work was supported by the National Natural Science Foundation of China (NSFC) via grant 61976072, 61632011 and 61772153.

\bibliography{emnlp-ijcnlp-2019}
\bibliographystyle{acl_natbib}

\onecolumn
\appendix

\section{Appendices for ``Cross-Lingual BERT Transformation for Zero-Shot Dependency Parsing"}
\label{sec:appendix}

\subsection{Statistics of UD (v2.2) Treebanks}
\label{sec:ud-details}
The statistics of the Universal Dependency treebanks we used are summarized in Table~\ref{tbl:ud-statistics}.

\begin{table*}[hpbt]
	\centering
	\begin{tabular}{l|l|c|c}
		\hline
		Language & Language Family & Treebank & Test Sentences  \\
		\hline
		English (en) &IE.Germanic &EWT &2,077 \\
		\hdashline
		German (de) &IE.Germanic &GSD &977 \\
		Danish (da) & IE.Germanic &DDT &565 \\
		Swedish (sv) &IE.Germanic &Talbanken &1,219 \\
		Dutch (nl) &IE.Germanic &Alpino, LassySmall &1,472 \\
		\hdashline
		French (fr) &IE.Romance &GSD &416 \\
		Italian (it) &IE.Romance &ISDT &482 \\
		Spanish (es) &IE.Romance &GSD, AnCora &2,147 \\
		Portuguese (pt) &IE.Romance &Bosque, GSD &1,681 \\
		Romanian (ro) &IE.Romance &RRT &729 \\
		\hdashline
		Slovak (sk) &IE.Slavic &SNK &1,061 \\
		Polish (pl) &IE.Slavic &LFG, SZ &2,827 \\
		Bulgarian (bg) &IE.Slavic &BTB & 1,116 \\
		Slovenian (sl) &IE.Slavic &SSJ, SST &1,898 \\
		Czech (cs) &IE.Slavic &PDT, CAC, CLTT, FicTree &12,203 \\
		\hdashline
		Finnish (fi) &Uralic &TDT &1,555 \\
		Estonian (et) &Uralic &EDT &2,737 \\
		\hdashline
		Latvian (lv) &IE.Baltic &LVTB &1,228 \\
		\hline
	\end{tabular}
	\caption{Statistics of the Universal Dependeny treebanks we selected in our experiments. For language family, “IE” is the abbreviation for Indo-European. }
	\label{tbl:ud-statistics}
\end{table*}

\subsection{Implementation Details}

For the graph-based Biaffine parser, we exclude the learned embeddings in our re-implementation, to focus on the effect of pre-trained embeddings.
Besides, the universal POS tags are used throughout our experiments. 

The PyTorch version of the base BERT model for English and multi-languages\footnote{\url{github.com/huggingface/pytorch-pretrained-BERT}} are used to generate the 768-dimensional contextualized embeddings for English and target languages respectively. 
In the GD-based method, we use Adam optimizer, with a learning rate of 0.001, $\beta_1=0.9$, $\beta_2=0.999$.

\subsection{Full Results on UD Treebanks}
\label{sec:ud-uas}
The LAS of our models (including the combination of cross-lingual FastText embeddings and our CLBT ones, where they are concatenated as the input to the parser) and the baseline ones are shown in Table~\ref{tbl:main-results-las}, and UAS in Table~\ref{tbl:main-results-uas}.

\begin{table*}[htbp]
	\centering
	\small
	\renewcommand{\arraystretch}{1.2}
	\begin{tabular}{c|c|c|c|c||c|c}
		\hline
		\multirow{2}{*}{\bf Lan.}& Static & \multicolumn{5}{c}{Contextualized} \\
		\cline{2-7}
		&\bf FT-SVD &\bf mBERT &\bf CLBT (SVD) &\bf CLBT (SVD) +FT &\bf CLBT (GD) &\bf CLBT (GD) +FT\\
		\hline
		en &88.31 &90.71 &91.03 &\bf 91.32 &91.03 &\bf 91.32 \\
		\hdashline\hdashline
		de &59.31 &63.41 &64.47 &\bf 64.78 &62.14 &63.05 \\
		da &68.81 &70.57 &71.60 &\bf 72.03 &71.66 &71.57 \\
		sv &73.49 &70.09 &73.33 &75.70 &75.95 &\bf 76.72 \\
		nl &60.11 &65.66 &65.45 &\bf 65.90 &63.86 &64.92 \\
		\hdashline
		fr &73.46 &72.97 &74.70 &75.56 &\bf 76.59 &76.38 \\
		it &76.23 &79.02 &\bf 79.46 &79.18 &78.98 &79.27 \\
		es &66.91 &65.43 &67.14 &67.47 &\bf 68.33 &67.71 \\
		pt &67.98 &67.11 &69.12 &69.00 &\bf 69.25 &69.09 \\
		ro &52.11 &46.40 &55.14 &54.79 &\bf 55.84 &55.53 \\
		\hdashline
		sk &56.98 &50.76 &59.46 &59.43 &\bf 59.92 &59.60 \\
		pl &58.59 &63.10 &65.37 &65.71 &65.80 &\bf 66.80 \\
		bg &66.68 &\bf 71.20 &70.26 &70.33 &70.75 &70.89 \\
		sl &54.57 &56.78 &57.42 &57.36 &57.21 &\bf 57.68 \\
		cs &52.80 &45.20 &52.20 &52.37 &52.99 &53.05 \\
		\hdashline
		fi &48.74 &49.56 &51.00 &53.26 &52.61 &\bf 53.91 \\
		et &44.40 &46.64 &47.79 &48.27 &48.52 &\bf 48.57 \\
		\hdashline
		lv &49.59 &45.11 &48.59 &50.04 &49.78 &\bf 50.98 \\
		\hline
		AVG. &60.63 &60.53 &63.09 &63.60 &63.54 &\bf 63.87 \\
		
		\hline
	\end{tabular}
	\caption{Results (LAS\%) on the test sets. The two columns on the left show results of baseline models, while the others on the right show results of our models. Languages are split by language families with dashed lines. AVG. means the average of results from all target languages. (\textbf{Lan.} stands for Language, \textbf{FT} stands for FastText.) }
	\label{tbl:main-results-las}
	\vspace{-1em}
\end{table*}

\begin{table*}[htpb]
	\centering
	\small
	\renewcommand{\arraystretch}{1.2}
	\begin{tabular}{c|c|c|c|c||c|c}
		\hline
		\multirow{2}{*}{\bf Lan.}& Static & \multicolumn{5}{c}{Contextualized} \\
		\cline{2-7}
		&\bf FT-SVD &\bf mBERT &\bf CLBT (SVD) &\bf CLBT (SVD) +FT &\bf CLBT (GD) &\bf CLBT (GD) +FT\\
		\hline
		en &90.44 &92.49 &92.81 &\bf 93.11 &92.81 &\bf 93.11 \\
		\hdashline\hdashline
		de &69.49 &72.34 &\bf 73.72 &\bf 73.72 &71.08 &71.51 \\
		da &77.36 &79.29 &79.63 &\bf 80.05 &79.16 &79.70 \\
		sv &81.23 &78.25 &80.57 &82.28 &82.64 &\bf 83.34 \\
		nl &67.88 &73.22 &72.80 &\bf 73.30 &71.00 &72.11 \\
		\hdashline
		fr &78.35 &78.79 &80.01 &\bf 81.10 &80.85 &80.92 \\
		it &81.10 &83.73 &\bf 84.53 &84.22 &83.33 &83.95 \\
		es &74.92 &73.97 &75.52 &\bf 75.89 &75.70 &75.59 \\
		pt &76.46 &75.09 &\bf 77.17 &76.90 &76.71 &76.44 \\
		ro &63.23 &58.45 &66.01 &66.07 &\bf 66.30 &66.00 \\
		\hdashline
		sk &65.41 &60.19 &67.56 &\bf 68.31 &67.62 &67.83 \\
		pl &71.89 &74.03 &76.68 &76.25 &76.52 &\bf 77.04 \\
		bg &78.05 &\bf 82.83 &82.14 &82.01 &81.51 &81.70 \\
		sl &66.27 &67.86 &69.04 &\bf 69.16 &68.26 &68.59 \\
		cs &61.88 &54.86 &61.02 &61.29 &61.26 &61.26 \\
		\hdashline
		fi &66.36 &65.45 &65.65 &68.28 &67.96 &\bf 69.16 \\
		et &65.25 &64.22 &65.26 &65.87 &\bf 66.76 &66.49 \\
		\hdashline
		lv &71.43 &61.73 &65.54 &66.98 &67.41 &68.20 \\
		\hline
		AVG. &71.56 &70.84 &73.11 &73.63 &73.18 &73.52 \\
		\hline
	\end{tabular}
	\caption{Results (UAS\%) on the test sets. The two columns on the left show results of baseline models, while the others on the right show results of our models. AVG. means the average of results from all target languages. (\textbf{Lan.} stands for Language, \textbf{FT} stands for FastText.) }
	\label{tbl:main-results-uas}
	\vspace{-1em}
\end{table*}

\end{document}